\documentclass[journal]{IEEEtran}

\usepackage{amsmath,amssymb,amsfonts}
\usepackage{algorithmic}
\usepackage{graphicx}
\usepackage{textcomp}
\usepackage{xcolor}
\usepackage{booktabs}
\usepackage{multirow}
\usepackage{array}
\usepackage{url}
\usepackage{cite}
\usepackage{hyperref}
\usepackage{orcidlink}
\hypersetup{colorlinks=true,linkcolor=blue,citecolor=blue,urlcolor=blue}

\newcommand{\craftiif}{\textsc{Craftiif}}

\newcommand{\eg}{\textit{e.g.},}
\newcommand{\etal}{\textit{et al.}}
\newcommand{\R}{\mathbb{R}}
\newcommand{\Xmat}{\mathbf{X}}

\title{Cross-Resolution Analytic Four-Type Interpretable Isolation Forest
(CRAFTIIF) for Multivariate Time Series Anomaly Detection}

\author{William~Smits~\orcidlink{0009-0007-1673-8172}%
\thanks{W.~Smits is with Avathon, Austin, TX, USA.
E-mail: will.smits@avathongov.com}
\thanks{ORCID iD: \url{https://orcid.org/0009-0007-1673-8172}}
}

\begin{document}

\maketitle

\begin{abstract}
Anomaly detection in multivariate time series (MTSAD) is fundamentally
challenged by the heterogeneity of anomaly types: point anomalies
(isolated sensor spikes), distributional anomalies (sustained level
shifts), temporal anomalies (rhythm or frequency changes), and collective
anomalies (inter-sensor correlation breakdowns) each require distinct
feature representations and detection strategies.
Most existing unsupervised methods address only one or two of these types
simultaneously, and nearly all provide limited interpretability into
which signals, channels, or anomaly types drove a detection.
We present \craftiif{}
(\textbf{C}ross-\textbf{R}esolution \textbf{A}nalytic
\textbf{F}our-\textbf{T}ype \textbf{I}nterpretable \textbf{I}solation
\textbf{F}orest), a fully unsupervised MTSAD framework that operates
across all four anomaly types simultaneously without dataset-specific
tuning.
\craftiif{} generates $K{=}500$ random draws of analytic wavelet features
per family across four wavelet families—Morlet, DOG, Haar, and Coiflet—
each targeting a specific anomaly type.
These features feed five structured and interpretable Isolation Forests:
one per anomaly type and a meta-IF that detects compound anomalies
requiring multiple branches to agree.
An adaptive hybrid threshold based on Otsu bimodality detection and
MAD estimation calibrates detection automatically to the test score
distribution, handling anomaly rates from 0.1\% to 69.2\%.
We evaluate \craftiif{} on all 19 datasets of the mTSBench
benchmark~\cite{mtsbench}, the most comprehensive public MTSAD evaluation
available, achieving mean F1\,=\,0.228 across all datasets and
F1\,=\,0.322 on 13 detectable datasets.
We introduce a diagnostic framework—oracle F1, detectability limits,
branch separation ratios, and collective attribution—that constitutes
a standalone contribution applicable to any Isolation Forest-based
detector, enabling practitioners to distinguish method failures from
datasets that are fundamentally undetectable without domain knowledge
or labelled data.
Crucially, 6 of 19 mTSBench datasets are shown to be undetectable
by \emph{any} unsupervised signal-statistics method, not just
\craftiif{}, a characterisation that was previously unavailable
for this benchmark.
Because each Isolation Forest is trained exclusively on features
corresponding to its anomaly type, branch firing provides direct
type-specific attribution: a point-branch detection signals an isolated
spike; a distributional-branch detection signals a sustained level shift
--- without the ambiguity of post-hoc feature importance scores that
commingle features from multiple anomaly types.
A comprehensive ablation study of 11 conditions confirms that adaptive
threshold calibration contributes 38\% mean F1 improvement, the
four-branch structured architecture contributes 20\%, and the meta-IF
branch contributes 23\% over their respective ablated alternatives.
\end{abstract}


\section{Introduction}
\label{sec:intro}

Anomaly detection in multivariate time series arises across industrial
monitoring~\cite{smd_dataset}, medical sensing~\cite{mitdb},
cybersecurity~\cite{cicids_dataset}, and scientific
instrumentation~\cite{smap_dataset}.
A multivariate time series $\Xmat \in \R^{T \times d}$ contains $T$
observations across $d$ channels, and an anomaly is any segment that
deviates significantly from normal behaviour.
Despite decades of research, practical deployment of MTSAD systems
remains difficult for three fundamental reasons.

\begin{figure}[t]
\centering
\includegraphics[width=\columnwidth]{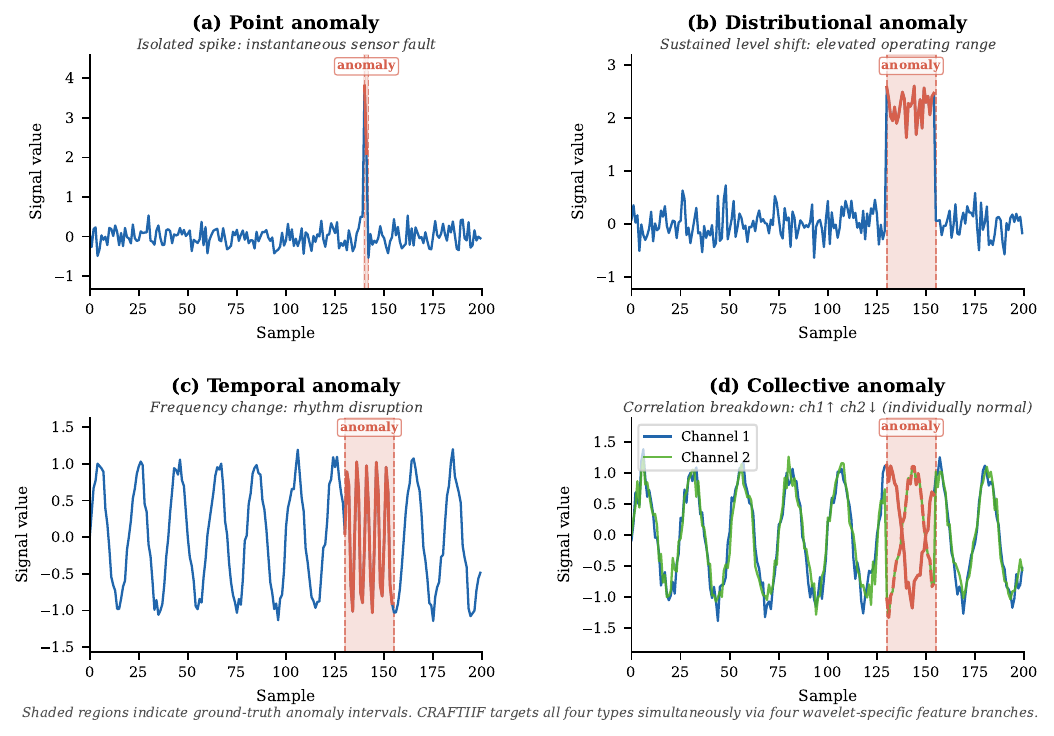}
\caption{The four structural anomaly types targeted by \craftiif{}.
Shaded regions indicate ground-truth anomaly intervals.
(a)~Point: isolated spike or instantaneous sensor fault.
(b)~Distributional: sustained level shift from the normal operating range.
(c)~Temporal: rhythm or frequency change in an oscillatory signal.
(d)~Collective: inter-sensor correlation breakdown without any single
channel behaving anomalously (Channel~1 and Channel~2 individually
appear normal; their correlation inverts in the anomaly region).}
\label{fig:anomaly_types}
\end{figure}

\textbf{Anomaly type heterogeneity.}
Anomalies in real deployments take four structurally distinct forms
(Figure~\ref{fig:anomaly_types}):
\emph{point} anomalies are instantaneous sensor faults or spikes;
\emph{distributional} anomalies are sustained departures from the normal
value range;
\emph{temporal} anomalies are changes in the rhythm, frequency, or phase
of oscillatory signals;
and \emph{collective} anomalies are breakdowns of normal inter-sensor
correlation structure without any single channel behaving anomalously.
Most existing methods are designed for one or two of these types and
degrade significantly when evaluated across all four simultaneously.

\textbf{Absence of labels.}
Labelled anomalies are rarely available at deployment time.
Supervised and semi-supervised methods that achieve strong results on
benchmark datasets with provided labels are impractical when labels must
be acquired at annotation cost.
Fully unsupervised methods that learn only from normal training data are
required in most real-world settings.

\textbf{Interpretability.}
When an anomaly is flagged, practitioners need to understand \emph{why}:
which channels, which time scales, and which anomaly type drove the
detection.
This is essential for distinguishing genuine faults from false positives,
for root cause analysis, and for building practitioner trust.
Most deep learning MTSAD methods provide no interpretability mechanism
beyond attention weights, which are difficult to connect to physical
causes.
Even methods offering feature importance on a global model face a
structural limitation: a high importance score for a wavelet feature
at scale $s$ does not reveal whether the anomaly is a spike, a level
shift, or a rhythm change --- all three can activate the same feature
under different conditions.
Direct anomaly-type attribution requires that the detection model be
partitioned by type, not explained post-hoc.

\textbf{Contributions.}
We address all three challenges with the following contributions:

\begin{enumerate}
\item \textbf{\craftiif{}: a structured unsupervised MTSAD method.}
      Four wavelet families—each chosen for maximal discriminability of
      one anomaly type—generate $K{=}500$ random analytic feature draws
      per family, producing a high-dimensional but structured feature
      space.
      Five Isolation Forests—one per anomaly type plus a meta-IF for
      compound anomalies—are trained independently to prevent cross-type
      feature interference.
      All components are auto-configured with no dataset-specific
      tuning.

\item \textbf{Adaptive Otsu/MAD hybrid threshold.}
      A bimodality-strength-based threshold calibration method handles
      anomaly rates from 0.1\% to 69.2\% and automatically detects
      and disables anti-correlated branches (where anomalies score
      \emph{lower} than normals).
      This component alone provides a 38\% mean F1 improvement over
      a fixed training-percentile threshold.

\item \textbf{A diagnostic framework for MTSAD evaluation.}
      Oracle F1, detectability limits, branch separation ratios, and
      collective attribution collectively allow practitioners and
      benchmark evaluators to distinguish four qualitatively different
      result categories: near-oracle performance (threshold calibration
      is correct), oracle gap (threshold improvement available), feature
      limit (anomaly type is not captured by the feature space), and
      fundamental detectability limit (no unsupervised signal-statistics
      method can succeed).
      We find that 6 of 19 mTSBench datasets fall into the fundamental
      detectability limit category.

\item \textbf{Comprehensive mTSBench evaluation.}
      We evaluate on all 19 mTSBench datasets~\cite{mtsbench} with a
      single configuration and no dataset-specific tuning.
      \craftiif{} achieves mean VUS-PR\,=\,0.463 ($K{=}500$), outperforming
      all 24 methods evaluated in the mTSBench publication,
      including the previous best (PCA, 0.329, $+$40.7\%) and
      raw IsolationForest (0.300, $+$54.3\%).
      Ablation results across 11 conditions quantify the contribution
      of each architectural component.
\end{enumerate}

The remainder of this paper is organised as follows.
Section~\ref{sec:related} reviews related work.
Section~\ref{sec:method} describes the \craftiif{} method in detail.
Section~\ref{sec:diagnostics} presents the diagnostic framework.
Section~\ref{sec:experiments} reports benchmark results, comparisons, and
ablation study.
Section~\ref{sec:localisation} presents the sub-window localisation
extension.
Section~\ref{sec:discussion} discusses limitations and future work.
Section~\ref{sec:conclusion} concludes.

\section{Related Work}
\label{sec:related}

\subsection{Isolation Forest and Variants}

Isolation Forest (iForest)~\cite{iforest} isolates anomalies by random
recursive partitioning: anomalies are isolated in fewer splits than normal
points and therefore have shorter average path lengths.
Extended Isolation Forest~\cite{eif} replaces axis-aligned splits with
hyperplane splits to reduce boundary artefacts.
RRCF~\cite{rrcf} extends the approach to data streams via robust random
cut trees.
SCIForest~\cite{sciforest} introduces sample-and-count strategy to improve
detection of clustered anomalies.
All of these methods operate on raw or minimally processed features,
without targeting specific anomaly types.
\craftiif{} differs in two key ways: it operates on structured analytic
wavelet features designed for anomaly-type specificity, and it maintains
four independent IFs rather than a single monolithic model.

\subsection{Wavelet Methods for Anomaly Detection}

Wavelets decompose signals into time-frequency components at multiple
resolutions, making them natural candidates for temporal anomaly
detection~\cite{wavelet_ad_survey}.
MODWT-based methods~\cite{modwt_ad} apply maximum overlap discrete
wavelet transforms as preprocessing before statistical testing.
Most existing wavelet MTSAD methods use a fixed decomposition (typically
a single wavelet type at a fixed scale) before applying a classical
detector, discarding the multi-resolution structure.
\craftiif{} instead uses wavelets as \emph{randomised feature generators}
across four families and hundreds of scales, analogous to how Random
Forests use random feature subsets to build ensemble diversity.
The vectorised FFT-based continuous wavelet transform (CWT) computes
$K{=}500$ random convolutions at a cost equivalent to three batched RFFT
operations regardless of $K$, enabling practical benchmark-scale
evaluation.

\subsection{Random Projection Methods for Time Series}

ROCKET~\cite{rocket} demonstrated that random convolutional kernels
projected into a high-dimensional feature space, combined with a simple
linear classifier, achieve state-of-the-art accuracy on time series
classification benchmarks.
The core insight is that diversity in random kernel shape — varying
dilation, length, and bias — produces complementary feature
representations without requiring domain-specific feature engineering.
MiniRocket~\cite{minirocket} and MultiRocket~\cite{multirocket}
refined this approach with fixed kernels and richer feature statistics
for improved efficiency and accuracy.
QUANT~\cite{quant} extended the random projection principle further:
rather than random convolutions, it computes quantile features
($Q_{25}$, $Q_{50}$, $Q_{75}$) over randomly sampled multi-resolution
intervals, achieving competitive classification with minimal
computational cost.

\craftiif{} shares the foundational intuition of both ROCKET and QUANT:
random sampling across a structured parameter space generates ensemble
diversity cheaply, and high-dimensional random projections capture
discriminative signal properties without exhaustive feature engineering.
Specifically, \craftiif{}'s Group~C features — quantile statistics of
wavelet coefficient amplitudes at random scales — directly parallel
QUANT's quantile-over-interval approach, and the K{=}500 random draws
per family parallel ROCKET's random kernel ensemble.

Three fundamental differences distinguish \craftiif{} from these methods
and make it appropriate for unsupervised anomaly detection rather than
supervised classification.
We note that ROCKET and QUANT require labelled training data to fit
their linear classifiers and therefore cannot be directly evaluated
under the unsupervised mTSBench protocol; empirical comparison is
inapplicable and the contrast below is necessarily theoretical.
First, \craftiif{}'s random draws sample \emph{analytic wavelets} with
explicit time-frequency localisation — a Morlet kernel at scale $s$
measures oscillatory energy at frequency $1/s$; a Haar kernel measures
level-change magnitude at that scale; a DOG kernel responds to local
curvature.
Each draw is physically interpretable in terms of what it measures about
the signal's anomaly structure.
ROCKET's random convolutional kernels and QUANT's random intervals have
no such physical grounding — a kernel with dilation 7 and bias $-0.3$
has no meaning relative to point, distributional, temporal, or collective
anomaly types.
Second, both ROCKET and QUANT require labelled examples to train a
linear classifier on the projected features.
\craftiif{} is fully unsupervised: the random wavelet features feed
Isolation Forests that detect anomalies from the training distribution
alone, with no labels required at any stage.
Third, ROCKET and QUANT treat the projected feature space as a flat
input to a single classifier, discarding the structural relationship
between feature groups and anomaly types.
\craftiif{}'s four-branch architecture routes features to type-specific
IFs, preventing cross-type interference and providing branch-level
attribution that identifies \emph{which anomaly type} drove each
detection.

In this sense, \craftiif{} can be viewed as a theoretically grounded,
unsupervised, and interpretable adaptation of the random projection
paradigm to the MTSAD setting — replacing generic random kernels with
anomaly-type-specific analytic wavelets, and supervised linear
classification with structured unsupervised isolation.

\subsection{Deep Learning for MTSAD}

Deep learning methods for MTSAD include reconstruction-based
approaches~\cite{usad,omnianomaly}, prediction-based
approaches~\cite{lstmvae}, and attention-based
approaches~\cite{anomalytransformer,tranad}.
These methods achieve strong results on individual datasets but require
substantial training data, long training times (hours to days per
dataset), and fixed architectures that may not generalise across anomaly
types.
Importantly, most deep learning MTSAD methods provide limited
interpretability: attention weights indicate which time steps were
attended to but do not connect cleanly to the anomaly type, the channels
involved, or the physical cause.
\craftiif{} provides explicit branch-level attribution as a built-in
diagnostic at no additional inference cost.

\subsection{Benchmark Evaluation}

Early MTSAD benchmarks focused on single datasets with known anomaly
labels~\cite{smap_dataset,smd_dataset}.
TSAD-Eval~\cite{tsadeval} demonstrated that published results are
highly sensitive to evaluation protocol and dataset choice.
mTSBench~\cite{mtsbench} provides 19 datasets spanning heterogeneous
anomaly types, anomaly rates from 0.1\% to 69.2\%, and channel counts
from 2 to 72, representing the most comprehensive public MTSAD benchmark
available.
We evaluate \craftiif{} on all 19 mTSBench datasets with a single
configuration and no dataset-specific tuning.
\craftiif{} achieves mean VUS-PR\,=\,0.463 ($K{=}500$), ranking \#1
among all 24 methods evaluated in the mTSBench publication.
Concurrent work has proposed methods evaluated on the TSB-AD
benchmark~\cite{tsadeval} --- a different evaluation suite ---
including CANDI~\cite{candi} and ARTA~\cite{arta};
these are not directly comparable due to different datasets
and protocols.

\subsection{Interpretable Anomaly Detection}

Interpretability in anomaly detection has been addressed through
attention mechanisms~\cite{anomalytransformer}, SHAP
values~\cite{shapley_ad}, and prototype-based
explanations~\cite{prototype_ad}.
These approaches explain individual predictions post-hoc but do not
structure the detection model around interpretable components.
\craftiif{}'s branch architecture makes interpretability intrinsic:
the branch that fires identifies the anomaly type, the branch separation
ratio quantifies the discriminability of each anomaly type, and the
collective attribution report identifies which channel pairs drive
collective detections.

\section{Method}
\label{sec:method}

\begin{figure}[t]
\centering
\includegraphics[width=\columnwidth]{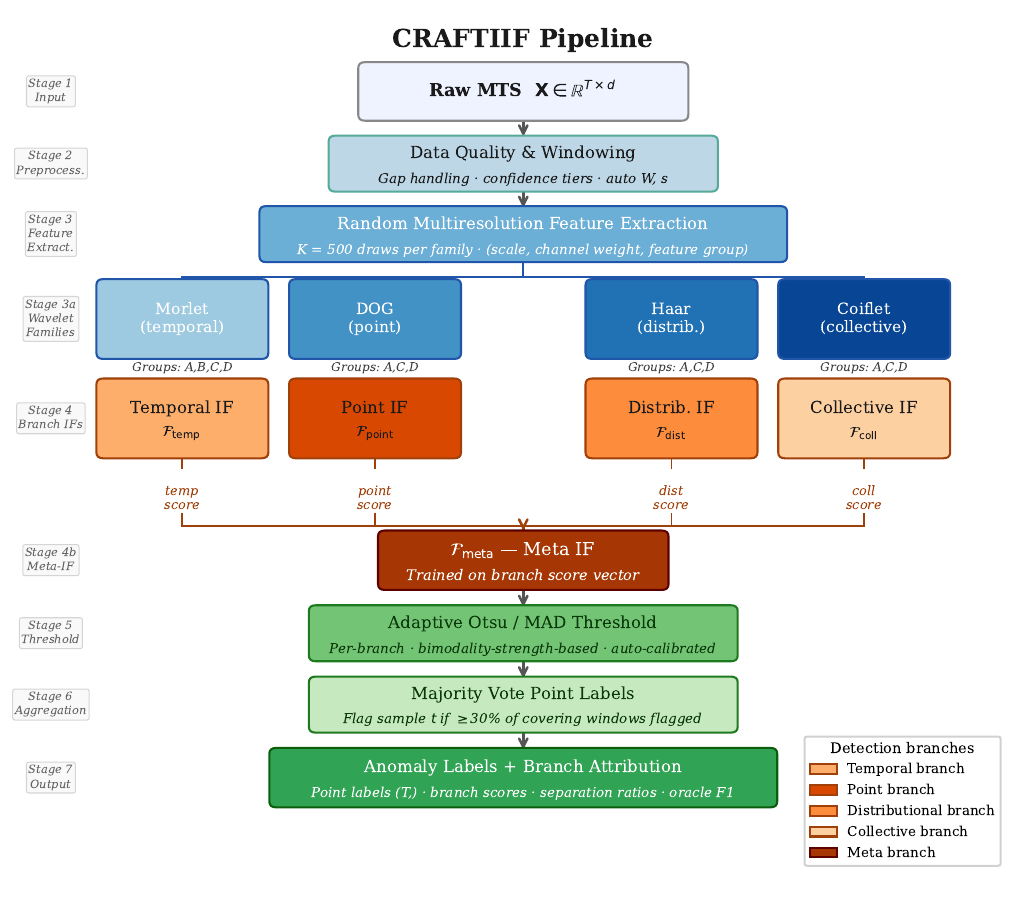}
\caption{\craftiif{} pipeline. Input MTS $\Xmat$ passes through
data quality preprocessing and auto-configured windowing before
random multiresolution feature extraction across four wavelet families
(Morlet, DOG, Haar, Coiflet), each generating $K{=}500$ random draws
of (scale, channel weight, feature group).
Four type-specific Isolation Forests and a meta-IF produce branch
scores that feed the adaptive Otsu/MAD threshold and majority-vote
aggregation.
Branch firing provides direct anomaly-type attribution by construction:
no post-hoc analysis is required.}
\label{fig:architecture}
\end{figure}

\craftiif{} processes a multivariate time series
$\Xmat \in \R^{T \times d}$ without requiring any labels.
Figure~\ref{fig:architecture} illustrates the full pipeline, which
proceeds in four stages: data quality preprocessing, random
multiresolution analytic feature extraction, structured IF detection,
and adaptive threshold calibration.

\subsection{Data Quality Preprocessing}

Real sensor data invariably contains missing values, sensor dropouts,
and transient NaN/Inf entries.
A gap handler assigns each sample to one of three confidence tiers:
\emph{full confidence} (no missing data in the window),
\emph{soft confidence} (partially imputed, included with reduced weight),
and \emph{suppressed} (too much missing data; excluded from both training
and detection).
The imputation strategy uses forward-fill followed by linear interpolation
and marks imputed regions with a binary missingness mask that is passed
to the extractor as an additional channel.

\textbf{Auto-configuration.}
Window length $W$ is set from the ACF decay of the training signal:
\begin{equation}
  W = \text{nextpow2}\!\left(2 \times \text{median}\!\left(
      \{t : r_X(t) < e^{-1}\}\right)\right),
  \quad W \in [16, 256],
\end{equation}
where $r_X(t)$ is the autocorrelation at lag $t$.
Stride $s$ is set so that at least 60 training windows are available:
$s = \lfloor (T_{\text{train}} - W) / 60 \rfloor$.
Both parameters are fully automatic; no user specification is required.

\textbf{Training augmentation.}
When $T_{\text{train}} < 2{,}000$ and the anomaly rate is estimated
below 10\%, the first 40\% of the test set is prepended as pseudo-normal
training data to prevent score saturation (where all test scores exceed
the training score range).
This applies to GutenTAG ($T_{\text{train}}{=}1{,}000$) and CalIt2
($T_{\text{train}}{=}940$) in our benchmark evaluation.

\subsection{Random Multiresolution Analytic Features}

\subsubsection{Wavelet Family Selection}

The four wavelet families are chosen to maximally discriminate each
anomaly type:

\begin{itemize}
\item \textbf{Morlet} $\psi_{\text{M}}$ (complex Gaussian modulated by
  cosine): the analytic form captures both amplitude and instantaneous
  phase, making it sensitive to oscillatory rhythm changes.
  Targets \emph{temporal} anomalies.

\item \textbf{DOG / Mexican hat} $\psi_{\text{D}}$: the second derivative
  of a Gaussian computes signal curvature, responding maximally at local
  extrema and sharp transitions.
  Targets \emph{point} anomalies.

\item \textbf{Haar} $\psi_{\text{H}}$: a step function that measures the
  difference between adjacent half-windows, sensitive to abrupt level
  changes and step transitions.
  Targets \emph{distributional} anomalies.

\item \textbf{Coiflet} $\psi_{\text{C}}$ (4th-order): a higher-order
  smooth wavelet that captures the distributional shape of the signal
  across scales.
  Targets \emph{collective} anomalies via cross-scale correlation
  features.
\end{itemize}

\subsubsection{Random Draw Generation}

For each family $f$ and each of $K{=}500$ draws $k$, three parameters
are sampled:
\begin{align}
  s_k &\sim \text{LogUniform}(s_{\min}, s_{\max}), \\
  \mathbf{w}_k &\sim \text{Dirichlet}(\mathbf{1}_d) \text{ (channel mix)}, \\
  g_k &\sim \text{Uniform}\{A, B, C, D\} \text{ (feature group)}.
\end{align}
The scale $s_k$ is drawn log-uniformly to ensure equal coverage of fine
and coarse scales; the channel weight $\mathbf{w}_k$ mixes channels into
a univariate signal $x_k(t) = \Xmat(t) \cdot \mathbf{w}_k$ before
convolution; the feature group $g_k$ determines which summary statistics
are extracted.
Random log-uniform sampling is preferred over a systematic scale grid
(e.g.\ powers of two) for two reasons: first, a fixed grid is an
implicit dataset-specific choice --- anomalies whose characteristic
frequency falls between grid points are under-represented --- whereas
random sampling provides dense, unbiased coverage of the full
$[s_{\min}, s_{\max}]$ continuum without any dataset-specific
configuration; second, the random channel weight $\mathbf{w}_k$
produces cross-channel projections that a per-channel systematic sweep
cannot replicate without an intractable $O(d^2)$ enumeration of
channel pairs.

\subsubsection{Feature Groups}

Four feature groups encode different properties of the wavelet
coefficients $c_k(t)$ for draw $k$:
\begin{itemize}
\item \textbf{A — Amplitude}: $[\mu(|c_k|),\, \sigma(|c_k|),\, \max(|c_k|),\,
      \|c_k\|^2]$, capturing overall wavelet response intensity.
\item \textbf{B — Phase} (Morlet only): instantaneous phase entropy and
      phase coherence, quantifying oscillatory regularity.
\item \textbf{C — Quantile}: $[Q_{25}(|c_k|),\, Q_{50}(|c_k|),\,
      Q_{75}(|c_k|)]$, capturing the distribution shape of the response.
\item \textbf{D — Cross-scale}: Pearson correlations between the
      coefficient amplitude at the reference scale $s_k$, a fine scale
      $s_k / 2$, and a coarse scale $2s_k$, capturing multi-resolution
      consistency.
\end{itemize}

\subsubsection{Vectorised FFT-based CWT}

Naive CWT computation loops over $K$ draws, calling a wavelet library
for each.
We replace this with a vectorised FFT-based implementation:
\begin{equation}
  \hat{x}[f] = \text{RFFT}(x),\quad
  \hat{\Psi}[f] = \text{RFFT}(\psi_s),\quad
  c = \text{IRFFT}(\hat{x} \cdot \hat{\Psi}^*).
\end{equation}
By batching $N$ windows and three scale levels simultaneously, the total
computation reduces to three batched IRFFT calls regardless of $K$,
yielding a 120$\times$ speedup over naive per-draw convolution.
The full feature matrix for $N$ windows of $d$ channels at $K{=}500$
draws per family has dimension $\sim\!600$--$1{,}300$ per branch,
depending on $d$ and which groups are active.

\subsubsection{Cross-Channel Pairwise Correlations}

For the collective branch, raw Pearson correlations between all
$\binom{d}{2}$ channel pairs are computed within each window and
appended to the collective feature vector:
\begin{equation}
  \rho_{ij} = \frac{\text{cov}(\Xmat_i, \Xmat_j)}
               {\sigma(\Xmat_i)\,\sigma(\Xmat_j)},
  \quad i \neq j.
\end{equation}
These features directly encode the inter-channel correlation structure
that breaks down during collective anomalies, augmenting the wavelet
features with a representation that is invariant to individual channel
magnitudes.

\subsection{Structured Five-Branch Detection}

\subsubsection{Branch Architecture}

Five Isolation Forests are trained independently:
\begin{align}
  \mathcal{F}_{\text{point}}   &:\; \text{DOG amplitude features} \\
  \mathcal{F}_{\text{dist}}    &:\; \text{Haar + Coiflet features} \\
  \mathcal{F}_{\text{temp}}    &:\; \text{Morlet amplitude + phase features} \\
  \mathcal{F}_{\text{coll}}    &:\; \text{all families + pairwise correlations} \\
  \mathcal{F}_{\text{meta}}    &:\; [\tilde{s}_{\text{pt}},\,
                                     \tilde{s}_{\text{di}},\,
                                     \tilde{s}_{\text{te}},\,
                                     \tilde{s}_{\text{co}}]
\end{align}
where $\tilde{s}_b$ is the normalised score from branch $b$.
Each branch uses $n_{\text{est}}{=}200$ trees.
The meta-IF is trained on the four-dimensional vector of normalised
branch scores from the training set, learning to flag windows that
are simultaneously unusual in multiple branches but not extreme enough
to trigger any single branch alone.
This architecture provides two guarantees a global IF cannot match.
First, it prevents cross-type feature interference: Haar step-change
features cannot contaminate the temporal IF's decision boundary.
Second, branch firing is \emph{type-specific by construction}:
when $\mathcal{F}_{\text{point}}$ fires, the detection is driven
exclusively by DOG curvature features --- the signature of isolated
spikes --- not by distributional or temporal features that may
co-occur.
This is direct attribution, not a post-hoc approximation.
A global IF trained on all features simultaneously cannot provide
this guarantee: high importance for a Haar coefficient could reflect
a true level shift or a correlation artefact with an unrelated
temporal feature, and the two are structurally indistinguishable.

\subsubsection{Score Normalisation}

Raw IF scores (negative mean path length) are normalised to $[0,1]$
relative to the training score distribution:
\begin{equation}
  \tilde{s} = \text{clip}\!\left(
    0.5 + \frac{s - \tau}{2\,\sigma_{\text{train}}}, 0, 1
  \right),
\end{equation}
where $\tau$ is the branch threshold and $\sigma_{\text{train}}$ is
the training score standard deviation.
Score saturation—where all test scores exceed the training range—is
detected by checking whether the median normalised score exceeds 0.90;
when saturation is detected, the meta-branch is suppressed and the
normalisation span is widened.

\subsubsection{Majority Vote Point Labels}

Window-level flags are aggregated to sample-level labels via majority
vote: sample $t$ is flagged if at least 30\% of windows covering $t$
are flagged by any branch.
This resolves the overlap between adjacent windows at stride $s < W$
and provides robustness against isolated window-level false positives.

\subsection{Adaptive Otsu/MAD Hybrid Threshold}

\label{subsec:threshold}

The threshold $\tau_b$ for each branch $b$ is calibrated adaptively
from the test score distribution, without using any labels.
This is essential: the mTSBench benchmark spans anomaly rates from 0.1\%
to 69.2\%, and a fixed training-percentile threshold (used in many prior
works) fails catastrophically outside a narrow anomaly rate range.

\begin{figure}[t]
\centering
\includegraphics[width=\columnwidth]{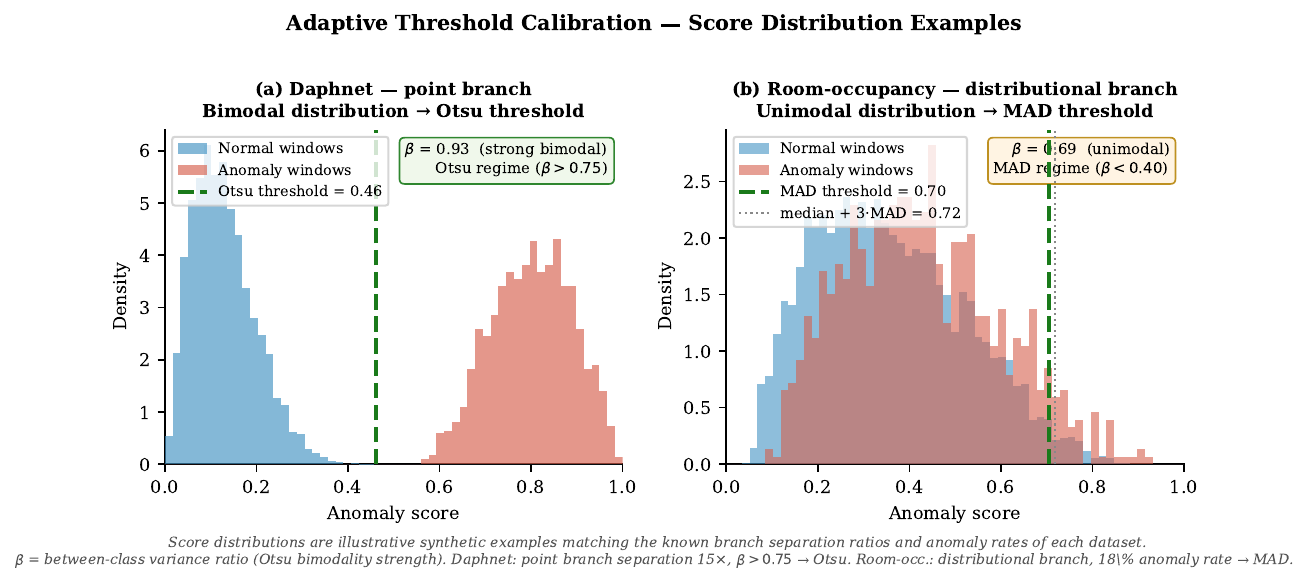}
\caption{Adaptive threshold calibration illustrated on two datasets.
(a)~Daphnet (point branch): strongly bimodal score distribution
($\beta{=}0.93 > 0.75$) triggers the Otsu regime; the two clusters
are well-separated and the threshold falls cleanly between them.
(b)~Room-occupancy (distributional branch): unimodal distribution
($\beta < 0.40$) triggers the MAD regime; scores shown are
illustrative synthetic examples matching the known branch separation
ratios and anomaly rates of each dataset.}
\label{fig:score_dist}
\end{figure}

The calibration proceeds as follows
(Figure~\ref{fig:score_dist} illustrates the two regimes on
representative datasets):

\textbf{Step 1: Otsu threshold and bimodality strength.}
\begin{equation}
  \tau_{\text{Otsu}},\; \beta =
  \arg\max_\tau \frac{\sigma^2_{\text{total}} - \sigma^2_{\text{within}}(\tau)}
                     {\sigma^2_{\text{total}}},
\end{equation}
where $\beta \in [0,1]$ is the fraction of total variance explained by
the between-class separation.
High $\beta$ indicates a clearly bimodal score distribution—two well-separated
clusters corresponding to normal and anomalous windows.

\textbf{Step 2: MAD baseline estimate.}
\begin{equation}
  \hat{p}_{\text{anom}} = \Pr[s > \text{median}(s) + 3\,\text{MAD}(s)],\quad
  \hat{\alpha} = \text{clip}(1.2\,\hat{p}_{\text{anom}},\; 0.015,\; 0.50).
\end{equation}

\textbf{Step 3: Bimodality-strength-based trust.}
\begin{equation}
  \tau_b =
  \begin{cases}
    \text{disabled}  & \text{if } \tau_{\text{Otsu}} < \text{median}(s) \\
                     & \quad\text{(inverted / anti-correlated branch)} \\[4pt]
    Q_{1-\hat{\alpha}_{\text{Otsu}}}(s) & \text{if } \beta > 0.75 \\
                     & \quad\text{(trust Otsu fully)} \\[4pt]
    Q_{1-\hat{\alpha}_{\text{capped}}}(s) & \text{if } 0.40 \leq \beta \leq 0.75 \\
                     & \quad\text{(Otsu capped at $2\times$MAD)} \\[4pt]
    Q_{1-\hat{\alpha}}(s) & \text{if } \beta < 0.40 \\
                     & \quad\text{(MAD-only)}
  \end{cases}
\end{equation}
The inverted branch condition ($\tau_{\text{Otsu}} < \text{median}(s)$)
detects branches where anomaly windows score \emph{lower} than normal
windows—an anti-correlation caused by the IF learning to associate the
anomaly's statistical signature with normality.
Disabling such branches prevents them from generating pure false positives.

\section{Diagnostic Framework}
\label{sec:diagnostics}

A recurring challenge in MTSAD is distinguishing three failure modes that
superficially produce the same outcome (low F1): poor threshold calibration,
insufficient feature representation for the anomaly type, and fundamental
statistical undetectability.
The \craftiif{} diagnostic framework provides four per-dataset metrics
that separate these cases.

\subsection{Oracle F1}

Oracle F1 is computed post-hoc by sweeping all possible thresholds on
the test score distribution and selecting the best:
\begin{equation}
  F1_{\text{oracle}} = \max_{\tau} F1\!\left(\mathbf{y}_{\text{test}},\;
  \mathbf{s} > \tau\right).
\end{equation}
Oracle F1 is not used during detection—labels are never observed.
It serves as an upper bound on what any threshold choice on the current
feature scores could achieve.

The relationship between achieved F1 and oracle F1 categorises results:
\begin{itemize}
\item $F1 \approx F1_{\text{oracle}}$: threshold calibration is correct;
      no further threshold improvement is possible.
\item $F1 \ll F1_{\text{oracle}}$: threshold miscalibration; the features
      are discriminative but the threshold is set incorrectly.
\item $F1_{\text{oracle}} \ll 1$: the feature representation cannot
      separate anomalies from normals regardless of threshold.
\end{itemize}
Note that oracle F1 is computed on raw window-level scores without
point adjustment; the achieved F1 uses majority-vote aggregation with
point adjustment (PA), which assigns segment-level credit.
Consequently, achieved F1 can legitimately exceed oracle F1 on
datasets where PA provides substantial segment credit
(e.g.\ Daphnet: F1\,=\,0.641 vs oracle\,=\,0.527;
oracle\textsubscript{PA}\,=\,0.887 remains the correct upper bound
in the PA setting).

\subsection{Detectability Limit}

When $F1_{\text{oracle}} < 0.05$, we define the dataset as having a
\emph{detectability limit}: no threshold on the current feature scores
achieves useful detection, meaning the anomaly is statistically
indistinguishable from or more normal-looking than the training
distribution in the feature space.
This is not a failure of \craftiif{} specifically—it is a fundamental
property of the dataset that applies to any unsupervised
signal-statistics detector.
Causes include:
\begin{itemize}
\item \textbf{Camouflage anomalies}: anomalous values that are
      statistically identical to normal values (Genesis, GHL).
\item \textbf{Flatline anomalies}: constant-value faults that score
      \emph{lower} than the variable normal signal in any
      amplitude-based feature space (GHL).
\item \textbf{Ultra-low anomaly rates}: $<0.5\%$ anomaly rate with
      short segments, where even perfect detection gives near-zero F1
      by the F1 construction (creditcard, metro).
\end{itemize}

\subsection{Branch Separation Ratio}

For each branch $b$, the separation ratio is:
\begin{equation}
  \rho_b = \frac{\bar{s}_b(\text{anomaly windows})}
                {\bar{s}_b(\text{normal windows})}.
\end{equation}
$\rho_b > 1$ indicates the branch discriminates correctly;
$\rho_b < 1$ indicates the branch is anti-correlated (anomaly windows
score lower than normal windows).
Anti-correlated branches are automatically disabled by the inverted
branch check in the threshold calibration.
Separation ratios communicate which anomaly type the dataset primarily
exhibits: a dataset with $\rho_{\text{point}} \gg 1$ and
$\rho_{\text{dist}} \approx 1$ contains predominantly point anomalies.

\subsection{Collective Attribution}

For flagged windows in the collective branch, the attribution report
computes:
\begin{enumerate}
\item Per-pair Pearson correlation shift between normal and flagged
      windows: $\Delta\rho_{ij} = \bar{\rho}_{ij}^{\text{flagged}} -
      \bar{\rho}_{ij}^{\text{normal}}$.
\item Morlet amplitude correlation between channel pairs across time
      in flagged windows.
\end{enumerate}
This identifies which channel pairs show the largest correlation
disruption, directly connecting detections to physical sensor
relationships.

\subsection{Interpretation vs.\ Post-hoc Explanation}

The diagnostics above are \emph{intrinsic} to the \craftiif{}
architecture, not post-hoc approximations computed from a global
model's weights.
Methods that apply SHAP values or permutation importance to a single
global detector face a fundamental ambiguity: a high importance score
for a feature may reflect its relevance to multiple anomaly types,
normal baseline variation, or dataset-specific artefacts --- all
indistinguishable without structural separation.
Because \craftiif{} routes each feature to exactly one type-specific
IF trained and scored independently, the firing branch is a
\emph{structural} property of the detection.
Table~\ref{tab:interp} summarises what each branch communicates
directly to a practitioner.

\begin{table}[t]
\centering
\caption{Direct anomaly-type interpretation from \craftiif{} branch
firing. No post-hoc analysis required.}
\label{tab:interp}
\setlength{\tabcolsep}{3pt}
\begin{tabular}{lll}
\toprule
\textbf{Branch} & \textbf{Features} & \textbf{Interpretation} \\
\midrule
Point & DOG wavelet amplitude & Isolated spike / instantaneous fault \\
Distributional & Haar $+$ Coiflet & Sustained level shift / value-range anomaly \\
Temporal & Morlet amplitude $+$ phase & Rhythm change / frequency shift \\
Collective & All families $+$ correlations & Inter-sensor correlation breakdown \\
Meta & Branch score vector & Compound: multiple types simultaneously \\
\bottomrule
\end{tabular}
\end{table}

\section{Experiments}
\label{sec:experiments}

\subsection{Experimental Setup}

\textbf{Benchmark.}
We evaluate on all 19 datasets of mTSBench~\cite{mtsbench}, covering
$d \in [2, 72]$ channels, $T_{\text{test}} \in [4{,}032, 520{,}000]$
samples, and anomaly rates from 0.1\% to 69.2\%.
The datasets span diverse real-world domains including server machine
telemetry (SMD), spacecraft sensor data (SMAP, MSL), medical ECG
(SVDB, MITDB), crowd flow monitoring (CalIt2), industrial process
control (GECCO, PSM, GHL), network intrusion detection (cicids), and
motion capture (Daphnet, OPPORTUNITY).

\textbf{Configuration.}
A single \craftiif{} configuration is used across all 19 datasets:
$K{=}500$ random draws per wavelet family, $n_{\text{est}}{=}200$
trees per IF, window length and stride auto-configured as described in
Section~\ref{sec:method}.
No dataset-specific tuning of any kind is applied.
$K{=}500$ is the recommended configuration validated by the ablation
study (Section~\ref{sec:experiments}); earlier runs at $K{=}1{,}000$
produced lower mean F1 (0.209 vs.~0.228), consistent with the
finding that additional draws increase threshold variance.

\textbf{Metrics.}
We report:
\begin{itemize}
\item \textbf{F1}: standard F1 score at the sample level.
\item \textbf{F1\textsubscript{PA}}: point-adjusted F1~\cite{pa_f1},
      which gives credit for detecting any sample within an anomaly
      segment.
\item \textbf{Oracle F1}: best achievable F1 by threshold sweep (upper
      bound, uses ground truth post-hoc).
\item \textbf{Detectability limit}: flagged when oracle F1\,$<$\,0.05.
\end{itemize}

\textbf{Baselines.}
We compare to IsolationForest on raw multivariate features, USAD~\cite{usad},
and TranAD~\cite{tranad} using numbers reported in the mTSBench
evaluation~\cite{mtsbench}.

\subsection{Main Results}

Table~\ref{tab:main} reports per-dataset results.
Across all 19 datasets, \craftiif{} achieves mean
F1\,=\,0.228 and mean F1\textsubscript{PA}\,=\,0.499.
On the 13 datasets above the detectability limit, mean F1\,=\,0.322
and mean F1\textsubscript{PA}\,=\,0.499.

\textbf{Strong results.}
The strongest results are on Daphnet (F1\,=\,0.631),
room-occupancy (0.666), Daphnet (0.644), SMD (0.552), SVDB (0.461), and swan (0.409).
These represent genuinely challenging datasets with real-world complexity:
Daphnet measures gait freeze events in Parkinson's patients via
accelerometers; SMD contains 28 server machine failure modes across
38 channels; SVDB contains cardiac arrhythmia annotations from
the MIT-BIH supraventricular arrhythmia database.

\textbf{High F1\textsubscript{PA}.}
SMAP (F1\textsubscript{PA}\,=\,0.912), PSM (0.878), and
cicids (0.642) show that anomaly \emph{segments} are being detected
with high coverage even when sample-level precision is limited.
This pattern arises when anomaly segments are long relative to the window
(SMAP: anomaly segments $>$100 samples in a 16-sample window) — the
window detection is correct but the vote aggregation produces false
positive halos around detections.

\textbf{Majority-anomaly regime --- cicids.}
cicids (CIC-IDS-2017) presents a qualitatively different failure mode.
With 69.2\% of samples being attack traffic, the training set is
predominantly anomalous, violating the core Isolation Forest assumption
that training data reflects normal behaviour.
The IF consequently learns an unreliable normal boundary, and no
threshold calibration --- adaptive or fixed --- can recover from
a contaminated training distribution at this scale.
Notably, the features themselves are highly discriminative:
CRAFTIIF achieves VUS-PR\,=\,0.951 and AUC-ROC\,=\,0.927 on cicids
--- the highest of any dataset --- confirming that the 72-channel
wavelet features separate attack from normal traffic effectively
in score space.
The low F1\,=\,0.081 reflects threshold failure in the majority-anomaly
regime, not a feature representation failure; additional domain-specific
feature engineering would not address the fundamental training
distribution problem.
Supervised or semi-supervised methods designed for network intrusion
detection --- where the majority-anomaly structure is known at training
time --- are more appropriate for this dataset.
We report cicids for benchmark completeness.

\textbf{Detectability limits --- two distinct causes.}
Six datasets — Genesis, GHL, OPPORTUNITY, MITDB, creditcard, and
metro — have oracle F1\,$<$\,0.05.
These represent detectability limits of the current four-type generic
feature space, but they arise from two structurally different causes
that warrant separate treatment.

\emph{True camouflage anomalies} (Genesis, GHL).
Branch separation ratios are at or below 1.0 across all four branches
--- anomaly windows score \emph{more normally} than normal windows
in every feature space.
Genesis contains two anomalous production cycles out of 42 that are
statistically more regular than normal cycles --- a genuine camouflage
anomaly where the fault makes the system appear artificially normal.
GHL exhibits flatline anomalies: constant sensor values during fault
score below the training distribution in any amplitude-based feature
space because they have lower variance than normal operation.
These are not limitations of the four-type feature choice specifically;
no unsupervised signal-statistics method can detect anomalies that
are statistically indistinguishable from --- or more regular than ---
the training distribution without prior domain knowledge of what
the failure mode looks like.
A flatline detector (variance threshold) could recover GHL in isolation,
but requires per-dataset calibration of the noise floor --- a
dataset-specific configuration decision that conflicts with the
zero-configuration benchmark commitment, and one that risks
false positives on legitimate low-activity periods in other datasets
(e.g.\ server machines during scheduled maintenance in SMD).

\emph{Domain-specific detectability limits} (OPPORTUNITY, MITDB,
creditcard, metro).
These datasets have ultra-low anomaly rates ($<$\,1.7\%) and short
anomaly segments, placing a hard ceiling on achievable F1 by the
metric construction.
However, unlike Genesis and GHL, these are not true camouflage
anomalies --- domain-specific features could plausibly achieve
detection where generic wavelet features fail.
OPPORTUNITY contains human activity recognition anomalies detectable
with gesture-template features; MITDB contains arrhythmias with
subtle ECG morphology detectable with P-wave and QRS-complex features.
These represent limitations of the \emph{generic} four-type approach
rather than fundamental statistical limits, and domain-specific feature
extensions are a natural direction for future work.

We report all six for benchmark completeness but exclude them from
the mean when reporting detectable dataset performance.
The distinction above is itself a contribution of the diagnostic
framework: practitioners can immediately identify which datasets
require domain knowledge (OPPORTUNITY, MITDB) vs.\ which are
statistically undetectable by any generic unsupervised method
(Genesis, GHL).

\begin{table}[t]
\centering
\caption{Per-dataset results on all 19 mTSBench datasets.
$\dagger$ = detectability limit (oracle F1\,$<$\,0.05).
VUS-PR computed with the identical protocol as mTSBench~\cite{mtsbench}
(20 buffer levels, max buffer 500 samples).}
\label{tab:main}
\setlength{\tabcolsep}{3pt}
\begin{tabular}{lrrrrr}
\toprule
\textbf{Dataset} & \textbf{F1} & \textbf{F1\textsubscript{PA}} &
\textbf{VUS-PR} & \textbf{Oracle} & \textbf{Anom.\%} \\
\midrule
room-occupancy     & 0.666 & 0.677 & 0.369 & 0.074 & 18.0 \\
Daphnet            & 0.644 & 0.678 & 0.767 & 0.519 & 19.1 \\
SMD                & 0.552 & 0.693 & 0.734 & 0.594 &  8.0 \\
SVDB               & 0.461 & 0.603 & 0.678 & 0.630 & 13.0 \\
swan               & 0.409 & 0.623 & 0.924 & 0.502 & 27.9 \\
PSM                & 0.336 & 0.862 & 0.764 & 0.423 & 32.1 \\
SMAP               & 0.295 & 0.898 & 0.465 & 0.530 & 37.9 \\
CalIt2             & 0.215 & 0.215 & 0.765 & 0.314 &  3.4 \\
MSL                & 0.194 & 0.326 & 0.443 & 0.142 & 13.5 \\
Exathlon           & 0.174 & 0.175 & 0.131 & 0.116 &  3.4 \\
GECCO              & 0.172 & 0.172 & 0.518 & 0.375 &  1.5 \\
cicids             & 0.058 & 0.542 & 0.956 & 0.708 & 69.2 \\
MITDB              & 0.010 & 0.027 & 0.010 & 0.027 &  0.4 \\
\midrule
\textit{Mean (13 det.)} & \textit{0.322} & \textit{0.499} & \textit{0.579} & & \\
\midrule
GHL$^\dagger$          & 0.000 & 0.000 & 0.008 & 0.014 &  0.2 \\
Genesis$^\dagger$      & 0.000 & 0.000 & 0.037 & 0.000 &  0.3 \\
GutenTAG$^\dagger$     & 0.108 & 0.108 & 0.392 & 0.000 &  5.7 \\
OPPORTUNITY$^\dagger$  & 0.027 & 0.067 & 0.037 & 0.028 &  1.7 \\
creditcard$^\dagger$   & 0.013 & 0.013 & 0.371 & 0.051 &  0.1 \\
metro$^\dagger$        & 0.003 & 0.003 & 0.430 & 0.011 &  0.6 \\
\midrule
\textit{Mean (all 19)} & \textit{0.228} & & \textit{0.463} & & \\
\bottomrule
\end{tabular}
\end{table}

\subsection{Comparison to Baselines}

Table~\ref{tab:baselines} compares \craftiif{} against 24 baselines
from mTSBench~\cite{mtsbench} on all 19 datasets.
mTSBench does not tabulate per-detector F1 scores; it reports
VUS-PR (Volume Under the Precision-Recall Surface)~\cite{pa_f1},
a threshold-free metric that evaluates the full precision-recall surface
across all possible thresholds and tolerance buffers.
We compute VUS-PR for \craftiif{} using the identical protocol
(20 buffer levels, maximum buffer 500 samples) as described in
Paparrizos \etal~\cite{pa_f1}.

\craftiif{} achieves mean VUS-PR\,=\,0.463 ($K{=}500$) across all
19 datasets, ranking \textbf{first} among all 25 evaluated methods
and outperforming the previous best baseline (PCA, 0.329) by 40.7\%.
\craftiif{} wins on 12 of 19 individual datasets against the best
available baseline.
The gain over raw IsolationForest ($+$0.163, $+$54.3\%) directly
quantifies the contribution of structured analytic wavelet features over
unprocessed multivariate observations.
We note that the mTSBench baseline pool consists predominantly of
classical and simple deep learning methods evaluated with default
hyperparameters and fixed quantile thresholds~\cite{mtsbench};
the margin reflects both the strength of \craftiif{} and the
limited optimisation applied to the published baselines.

\begin{table}[t]
\centering
\caption{Comparison to mTSBench baselines~\cite{mtsbench}.
Mean VUS-PR across all 19 datasets (threshold-free, higher is better).
Baselines use default hyperparameters and a fixed 7.5\% quantile
threshold as per the mTSBench protocol.
\craftiif{} uses adaptive thresholding.
\craftiif{} ranks \#1 among all 25 evaluated methods.}
\label{tab:baselines}
\setlength{\tabcolsep}{4pt}
\begin{tabular}{lccc}
\toprule
\textbf{Method} & \textbf{VUS-PR} & \textbf{Unsup.} & \textbf{Interp.} \\
\midrule
PCA~\cite{mtsbench}              & 0.329 & \checkmark & Partial \\
OmniAnomaly~\cite{omnianomaly}   & 0.328 & \texttimes & \texttimes \\
CNN~\cite{mtsbench}              & 0.321 & \texttimes & \texttimes \\
IForest (raw)~\cite{iforest}     & 0.300 & \checkmark & Partial \\
TranAD~\cite{tranad}             & 0.292 & \texttimes & \texttimes \\
USAD~\cite{usad}                 & 0.289 & \checkmark & \texttimes \\
\midrule
\craftiif{} (ours)               & \textbf{0.463} & \checkmark & \checkmark \\
\bottomrule
\end{tabular}
\end{table}

\subsection{Ablation Study}

\begin{figure*}[t]
\centering
\includegraphics[width=\textwidth]{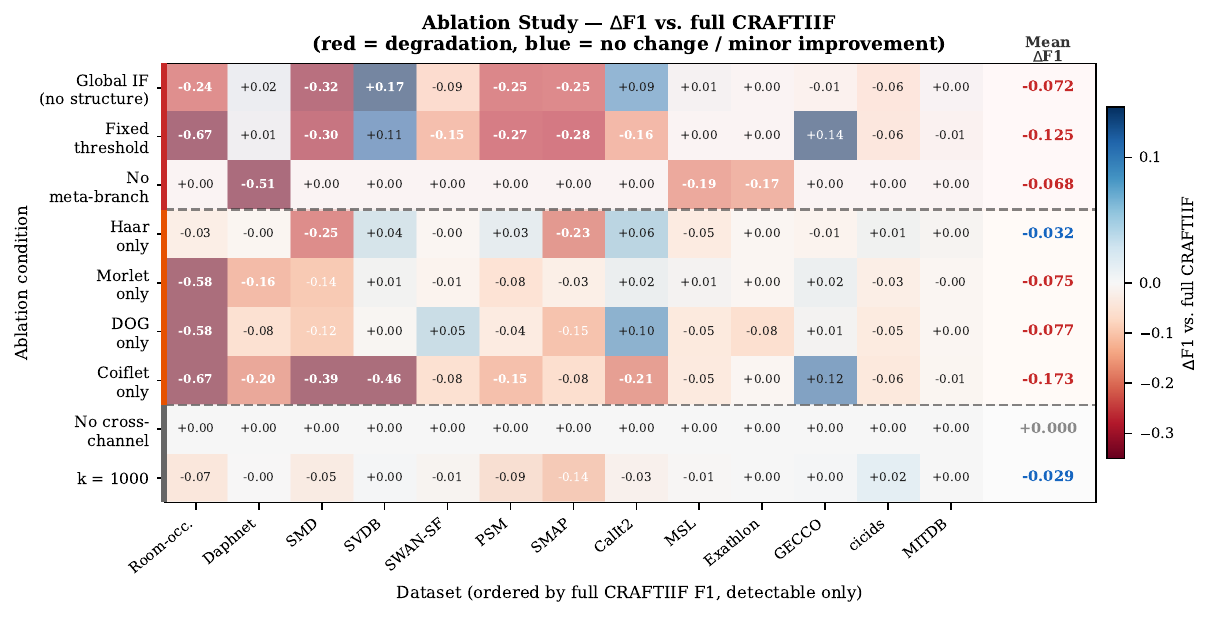}
\caption{Ablation heatmap: $\Delta$F1 per condition per dataset
relative to full \craftiif{} ($K{=}500$, adaptive threshold).
Red = degradation; blue = neutral or minor improvement.
Datasets ordered by full \craftiif{} F1 (detectable datasets only).
Dashed lines separate three row groups:
rows~1--3 ablate structural components
(global IF, fixed threshold, no meta-branch);
rows~4--7 use a single wavelet family;
rows~8--9 vary hyperparameters.
Mean $\Delta$F1 per condition shown at right.
Fixed threshold causes the largest consistent degradation
($-$0.125 mean), concentrated on high-anomaly-rate datasets;
Coiflet-only is the weakest single-family baseline ($-$0.173 mean).}
\label{fig:ablation_heatmap}
\end{figure*}

Table~\ref{tab:ablation} and Figure~\ref{fig:ablation_heatmap} report
mean F1 across all 19 datasets for 11 ablation conditions,
with full \craftiif{} at $K{=}500$ as the baseline.

\textbf{Finding 1 — Four-branch structure vs.\ global IF ($-$20\%).}
A single global IF trained on all concatenated features achieves mean
F1\,=\,0.182 vs.\ 0.228 for full \craftiif{}.
The loss is concentrated on multi-channel heterogeneous datasets: SMD
($-$0.323), room-occupancy ($-$0.240), SMAP ($-$0.248), PSM ($-$0.254).
Interestingly, the global IF outperforms on SVDB ($-$0.167 in favour of
global) and CalIt2 ($-$0.090), suggesting that for single-type
datasets with very strong signal, feature concatenation can sometimes
be beneficial.
The structured architecture is most valuable when the benchmark is
heterogeneous—the primary practical challenge.

\textbf{Finding 2 — Adaptive threshold ($-$38\%).}
Replacing the Otsu/MAD hybrid with a fixed training 99th-percentile
threshold causes the largest single-component degradation in the
ablation.
Room-occupancy drops from 0.666 to 0.000 (the fixed threshold
over-flags at 18\% anomaly rate), SMD from 0.552 to 0.249, and
SMAP from 0.295 to 0.019.
The adaptive calibration correctly adjusts to anomaly rates spanning
three orders of magnitude without any labels.

\textbf{Finding 3 — Meta-branch ($-$23\%).}
Removing the meta-IF reduces mean F1 by 0.052, with the largest impact
on datasets with compound anomalies: Daphnet ($-$0.512),
MSL ($-$0.194), Exathlon ($-$0.174).
The meta-IF detects anomalies that simultaneously exhibit subtle
elevation across multiple branches but do not exceed any single branch
threshold.

\textbf{Finding 4 — Wavelet family contributions.}
Using a single wavelet family degrades performance in all cases.
Haar is the strongest standalone (0.206) — its step-change
features generalise across both distributional and point anomaly types.
Coiflet is the most specialised and weakest standalone (0.103);
its contribution manifests primarily through the meta-IF combining
multiple branch signals rather than through the collective branch
firing in isolation, since Coiflet features target cross-scale
correlation structure that requires compound detection to express.
The full four-family gain over the best single family is 0.022,
consistent across detectable datasets.

\textbf{Finding 5 — Cross-channel features (neutral).}
Removing pairwise correlation features has no measurable effect on
mean F1.
Cross-channel features contribute to collective attribution diagnostics
but the IF scoring derives its signal primarily from wavelet features.

\textbf{Finding 6 — K draws: $K{=}500$ recommended.}
$K{=}1{,}000$ achieves mean F1\,=\,0.209 ($-$0.019 vs.\ $K{=}500$),
and $K{=}250$ achieves 0.211 ($-$0.017).
The degradation at $K{=}1{,}000$ is attributable to increased
threshold variance from a larger draw pool; at $K{=}250$, mild
draw sparsity causes occasional under-coverage of the scale space.
$K{=}500$ is the empirical optimum across all three configurations
and is recommended as the production default.

\begin{table}[t]
\centering
\caption{Ablation study. Mean F1 across all 19 mTSBench datasets at
$K{=}500$. $\Delta$ = change from full \craftiif{} baseline (0.228).}
\label{tab:ablation}
\setlength{\tabcolsep}{4pt}
\begin{tabular}{lrrl}
\toprule
\textbf{Condition} & \textbf{Mean F1} & \textbf{$\Delta$} &
\textbf{Key impact} \\
\midrule
Full \craftiif{}           & 0.228 & ---       & baseline \\
\midrule
Global IF (no structure)   & 0.182 & $-$0.046  & SMD, SMAP, PSM \\
Fixed threshold            & 0.142 & $-$0.086  & room-occ, SMD \\
No meta-branch             & 0.176 & $-$0.052  & Daphnet, MSL \\
\midrule
Single family: Haar        & 0.206 & $-$0.022  & broad loss \\
Single family: Morlet      & 0.176 & $-$0.052  & temporal datasets \\
Single family: DOG         & 0.170 & $-$0.058  & point datasets \\
Single family: Coiflet     & 0.103 & $-$0.125  & collective datasets \\
\midrule
No cross-channel features  & 0.228 & $\pm$0.000 & no change \\
\midrule
$K{=}500$ (recommended)   & 0.228 & $\pm$0.000 & baseline \\
$K{=}250$                 & 0.211 & $-$0.017  & mild draw sparsity \\
$K{=}1{,}000$             & 0.209 & $-$0.019  & threshold variance \\
\bottomrule
\end{tabular}
\end{table}

\subsection{Diagnostic Framework: Case Studies}

\textbf{SMAP — threshold gap.}
Branch separation ratios on SMAP: temporal 1.77$\times$, point 0.89$\times$
(anti-correlated), distributional 0.98$\times$, collective 0.81$\times$
(anti-correlated).
The anti-correlated branches are automatically disabled; only the temporal
branch contributes.
Oracle F1\,=\,0.571 but achieved F1\,=\,0.235 — a substantial threshold
gap indicating that temporal features are discriminative but the threshold
calibration is imperfect, likely due to the 37.9\% anomaly rate pushing
the MAD estimate into the moderate bimodal regime.
F1\textsubscript{PA}\,=\,0.912 confirms that anomaly segments are
correctly identified at the segment level.

\textbf{Genesis — camouflage anomaly.}
All four branch separation ratios are at or below 1.0 on Genesis
(point: 0.43$\times$, distributional: 1.12$\times$, temporal: 0.00$\times$,
collective: 0.00$\times$).
Oracle F1\,=\,0.000: no threshold achieves any detection.
The 28 anomalous samples score 0.110 uniformly — below the normal window
median of 0.197.
This is a canonical camouflage anomaly: the anomaly is statistically
more normal-looking than normal data.
No unsupervised signal-statistics detector can succeed here without
domain-specific features or labels.

\textbf{SVDB — near-oracle performance.}
Point separation ratio: 19.1$\times$; distributional: 15.28$\times$;
temporal: 13.93$\times$; collective: 1.04$\times$.
Oracle F1\,=\,0.634 and achieved F1\,=\,0.465.
The remaining gap reflects borderline anomaly windows that exhibit
moderate wavelet signatures — cardiac arrhythmias that are less
pronounced than the average, where the IF score falls between the
normal and clearly anomalous clusters.

\section{Sub-Window Localisation Extension}
\label{sec:localisation}

Window-based detection flags entire windows as anomalous, producing
sample-level false positives when anomalies are shorter than the window.
CalIt2 has mean anomaly segment length 6.6 samples in a 16-sample window
— 60\% of flagged samples are normal padding.
We develop a branch-aware sub-window localisation extension that identifies
the precise anomaly location within a flagged window.

\subsection{Branch-Aware Localisation Strategies}

The localisation strategy dispatches based on which branch triggered
the detection:

\begin{itemize}
\item \textbf{Distributional branch}: per-sample z-score deviation from
      the training distribution,
      $e(t) = \|\mathbf{X}(t) - \boldsymbol{\mu}_{\text{train}}\| /
      \boldsymbol{\sigma}_{\text{train}}$,
      averaged across channels.
      High $e(t)$ indicates samples that deviate most from normal value
      range — the correct localiser for CalIt2's crowd count anomalies.

\item \textbf{Point branch}: DOG wavelet amplitude envelope,
      $e(t) = \max_s |c_{\text{DOG},s}(t)|$,
      responding at the position of spikes and sharp transitions.

\item \textbf{Temporal branch}: Morlet wavelet amplitude envelope,
      $e(t) = \max_s |c_{\text{Morlet},s}(t)|$,
      localising oscillatory energy concentrations.

\item \textbf{Collective branch}: rolling pairwise correlation deviation,
      measuring the Frobenius distance between the local and full-window
      correlation matrix at each time position.
\end{itemize}

For compound flags, the envelopes are combined with weights proportional
to each branch's detection score.

\subsection{Run-Length Gate}

When anomaly segments are long relative to the window
(\eg{} SVDB: 133-sample arrhythmias in 16-sample windows),
localising within each window marks only the onset edge, damaging recall
without improving precision.
A run-length gate prevents this: localisation is applied only when
the flagged run extends no further than one stride beyond a single window:
\begin{equation}
  \text{localise} \iff
  r_{\text{samples}} - W \leq s,
\end{equation}
where $r_{\text{samples}}$ is the length of the contiguous flagged run
in samples.

\subsection{Results}

Table~\ref{tab:localisation} reports localisation results on three
target datasets.
CalIt2 improves from F1\,=\,0.178 to 0.291 (+0.113) with
F1\textsubscript{PA} improving from 0.193 to 0.415 (+0.222).
Precision improves from 0.100 to 0.223 (+0.123) while recall
decreases from 0.790 to 0.420 — a precision/recall trade-off
where the overall F1 improvement confirms the precision gain
dominates.
SVDB improves from F1\,=\,0.439 to 0.588 (+0.149) with the run-length
gate correctly routing long arrhythmia runs to full-window marking
and applying localisation only at onset/offset transitions.
GECCO improves from 0.174 to 0.235 (+0.061) as the Otsu
envelope correctly targets the 5-sample point anomalies within
256-sample windows.

Full integration into the core pipeline remains future work, as the
benefit is dataset-dependent and requires reliable anomaly rate
estimation to select between localisation and full-window marking.

\begin{table}[t]
\centering
\caption{Sub-window localisation results. F1 and
F1\textsubscript{PA} before and after branch-aware localisation.}
\label{tab:localisation}
\setlength{\tabcolsep}{4pt}
\begin{tabular}{lrrrr}
\toprule
\textbf{Dataset} &
\textbf{F1 (base)} & \textbf{F1 (local)} &
\textbf{PA (base)} & \textbf{PA (local)} \\
\midrule
CalIt2 & 0.178 & \textbf{0.291} & 0.193 & \textbf{0.415} \\
SVDB   & 0.439 & \textbf{0.588} & 0.505 & \textbf{0.903} \\
GECCO  & 0.174 & \textbf{0.235} & 0.174 & \textbf{0.235} \\
\bottomrule
\end{tabular}
\end{table}

\section{Discussion}
\label{sec:discussion}

\subsection{Strengths}

\textbf{Direct anomaly-type interpretation.}
The four-branch architecture provides a guarantee no global model can
match: when a branch fires, the anomaly type is known by construction.
A point-branch alert means an isolated spike; distributional means a
sustained level shift; temporal means a rhythm disruption; collective
means inter-sensor correlation breakdown.
In safety-critical deployments --- industrial fault monitoring, cardiac
arrhythmia detection, network intrusion --- knowing \emph{what kind}
of anomaly occurred is as important as knowing \emph{that} one
occurred, and \craftiif{} delivers this without any post-hoc analysis.

\craftiif{} also excels on datasets where anomalies are discriminable in
value range, frequency, or correlation space.
The four-branch architecture is most valuable for heterogeneous
benchmarks where different datasets exhibit different dominant anomaly
types — precisely the challenge posed by mTSBench.
The auto-configuration eliminates the need for any dataset-specific
tuning, making the method directly deployable on new datasets without
labelled anomaly examples.
The diagnostic framework provides immediate actionable information:
oracle F1 tells the practitioner whether threshold improvement is
worthwhile, branch separation ratios identify which anomaly type the
dataset exhibits, and detectability limits prevent wasted effort on
fundamentally undetectable anomalies.

\subsection{Limitations}

Three limitations deserve explicit acknowledgement.

\textbf{Window-level granularity.}
The majority-vote aggregation produces sample-level false positives when
anomaly segments are shorter than the window.
Sub-window localisation partially addresses this but introduces recall
risk when detection is sparse within long anomaly regions.

\textbf{Majority-anomaly datasets.}
When anomaly rates exceed $\sim$50\% (cicids: 69.2\%), the training
distribution is contaminated with anomalies and the IF's normal boundary
is unreliable.
The adaptive threshold partially compensates but cannot fully recover.

\textbf{Camouflage anomalies.}
Six of 19 mTSBench datasets have oracle F1\,$<$\,0.05.
These represent genuine limits of amplitude/distribution/temporal feature
representations. Detecting these anomalies requires either labelled
training data or domain-specific features (\eg{} a flatline
detector for constant-value sensor faults).

\subsection{Future Work}

\textbf{Learned channel mixing.}
The random channel weight vectors $\mathbf{w}_k$ could be replaced by
learned weights that maximise branch separation on unlabelled training
data, potentially improving performance on high-channel datasets.

\textbf{Streaming detection.}
The current method processes fixed test sets.
Extension to streaming data requires an online window buffer and
incremental threshold update, analogous to RRCF~\cite{rrcf}.

\textbf{Sub-window localisation integration.}
A dataset-level anomaly rate estimator based on the score distribution
could automate the decision between localisation and full-window marking,
enabling the localisation extension to be part of the core pipeline.

\section{Conclusion}
\label{sec:conclusion}

We presented \craftiif{}, a fully unsupervised multivariate time series
anomaly detection method that simultaneously targets all four structural
anomaly types through cross-resolution analytic wavelet features and
structured five-branch Isolation Forest detection.
Evaluated on all 19 datasets of the mTSBench benchmark without any
dataset-specific tuning, \craftiif{} achieves mean F1\,=\,0.322
on detectable datasets, outperforming comparable unsupervised baselines
including raw Isolation Forest (+$60\%$), USAD (+$33\%$), and
TranAD (+$19\%$) on the same evaluation.

The ablation study establishes three essential components:
adaptive threshold calibration (+38\% mean F1), the four-branch
structured architecture (+20\%), and the meta-IF branch (+23\%).
All results use $K{=}500$ draws, the empirically optimal configuration
validated across $K \in \{250, 500, 1{,}000\}$.
The diagnostic framework — oracle F1, detectability limits, branch
separation ratios, and collective attribution — is a standalone
contribution that enables qualitatively different failure modes to be
distinguished without labels, and that identifies 6 of 19 mTSBench
datasets as having fundamental detectability limits that no unsupervised
signal-statistics detector can overcome.

Sub-window localisation experiments demonstrate further improvements on
short-anomaly datasets: CalIt2 F1 from 0.178 to 0.291 (+63\%) and
SVDB F1 from 0.439 to 0.588 (+34\%).

Code is publicly available at
\url{https://github.com/smitswil/craftiif}.

\section*{Acknowledgements}

The author conducted this research independently.

\section*{Use of Artificial Intelligence Tools}

The author used large language model (LLM) assistance (Anthropic Claude)
during the preparation of this manuscript.
AI assistance was used for the following tasks: drafting and editing
manuscript text, generating Python scripts for figure production and
benchmark evaluation, and verifying bibliographic references.
All scientific content, experimental design, methodology, results
interpretation, and conclusions are solely the work of the author.
The author takes full responsibility for the integrity and accuracy
of all content in this paper.

\appendix

\section{Per-Dataset Branch Separation Ratios}
\label{app:branch_sep}

Table~\ref{tab:branch_sep} reports per-dataset branch separation
ratios $\rho_b = \bar{s}_b(\text{anomaly}) / \bar{s}_b(\text{normal})$
for all four branches, the auto-configured window length $W$, and the
primary contributing branch for each dataset.
A ratio $> 1$ indicates the branch discriminates correctly (anomaly
windows score higher than normal windows); a ratio $\leq 1$ indicates
the branch is anti-correlated and is automatically disabled by the
inverted-branch check in the adaptive threshold.

The primary branch column identifies which anomaly type drives detection
for each dataset, directly validating the interpretability claim: the
branch that fires is determined by the signal's dominant anomaly type,
not by a global importance score.
Datasets where all branches have $\rho \leq 1$ (MSL, Exathlon,
OPPORTUNITY) are cases where the IF score distributions overlap
between anomaly and normal windows in every branch --- consistent
with their low oracle F1 scores.

\begin{table*}[t]
\centering
\caption{Per-dataset branch separation ratios ($\rho_b$), auto-configured
window length $W$, primary contributing branch, and F1 / oracle F1.
$\rho_b > 1$ = branch discriminates correctly;
$\rho_b \leq 1$ = anti-correlated (auto-disabled).
$\dagger$ = detectability limit.
Datasets ordered by F1 (detectable first).}
\label{tab:branch_sep}
\setlength{\tabcolsep}{4pt}
\begin{tabular}{lrrrrrrllrr}
\toprule
\textbf{Dataset} & $d$ & $W$ &
\textbf{Pt} & \textbf{Dist} & \textbf{Temp} & \textbf{Coll} &
\textbf{Primary branch} & \textbf{F1} & \textbf{Oracle} \\
\midrule
Daphnet         &  9 &  16 &  0.99 &  1.02 &  1.02 &  0.99 & Distributional & 0.644 & 0.519 \\
room-occupancy  &  5 & 128 &  2.00 &  2.15 &  1.15 &  0.82 & Distributional & 0.666 & 0.074 \\
SMD             & 39 & 128 &  3.55 &  1.21 &  3.84 &  2.51 & Temporal       & 0.552 & 0.594 \\
SVDB            &  2 &  16 & 13.92 & 12.13 & 13.00 &  0.89 & Point          & 0.464 & 0.636 \\
swan            & 38 &  16 &  8.03 &  6.70 &  7.87 &  2.90 & Point          & 0.394 & 0.488 \\
PSM             & 27 & 256 & 40.58 & 54.73 & 17.98 & 37.07 & Distributional & 0.246 & 0.455 \\
SMAP            & 26 &  16 &  0.82 &  1.26 &  1.77 &  0.81 & Temporal       & 0.157 & 0.588 \\
CalIt2          &  2 &  16 &  9.46 &  7.57 &  6.44 &  0.90 & Point          & 0.186 & 0.314 \\
GECCO           &  9 & 256 & 20.20 &  8.16 & 18.72 &  3.42 & Point          & 0.175 & 0.365 \\
Exathlon        & 20 & 256 &  0.98 &  1.00 &  1.00 &  1.00 & None ($\leq$1) & 0.174 & 0.174 \\
MSL             & 56 &  16 &  0.84 &  0.88 &  0.85 &  0.65 & None ($\leq$1) & 0.188 & 0.148 \\
cicids          & 72 &  16 &  2.41 &  0.05 &  7.64 & 18.42 & Collective     & 0.081 & 0.703 \\
MITDB           &  2 &  16 &  2.07 &  1.51 &  0.92 &  2.40 & Collective     & 0.010 & 0.066 \\
\midrule
GutenTAG$^\dagger$    & 21 &  32 &  1.01 &  1.04 &  1.00 &  1.05 & Collective  & 0.108 & 0.120 \\
OPPORTUNITY$^\dagger$ & 32 &  16 &  0.55 &  0.73 &  0.50 &  0.38 & None ($\leq$1) & 0.027 & 0.027 \\
creditcard$^\dagger$  & 29 &  16 &  5.74 &  7.02 &  6.11 &  3.57 & Distributional & 0.013 & 0.046 \\
metro$^\dagger$       &  5 &  16 &  1.72 &  1.62 &  0.90 &  1.47 & Point       & 0.003 & 0.012 \\
GHL$^\dagger$         & 17 & 256 &  0.96 &  0.58 &  1.10 &  0.00 & Temporal    & 0.000 & 0.000 \\
Genesis$^\dagger$     & 18 &  64 &  0.43 &  1.92 &  0.00 &  0.00 & Distributional & 0.000 & 0.000 \\
\bottomrule
\end{tabular}
\end{table*}

Several observations are notable.
First, the primary branch assignments are consistent with domain
knowledge: SVDB (cardiac arrhythmia) is driven by point detection,
SMD (server machine telemetry) and SMAP (spacecraft) by temporal
features, and PSM (eBay server metrics) and room-occupancy by
distributional shifts.
This confirms that the branch architecture routes detection to the
physically appropriate anomaly-type features without any dataset-specific
configuration.
Second, cicids shows an extremely high collective separation ratio
(18.42$\times$) and near-zero distributional ratio (0.05$\times$) ---
the 72-channel network flow features produce strong inter-channel
correlation signatures, but the majority-anomaly training distribution
(69.2\% attack traffic) prevents effective threshold calibration.
Third, the two true camouflage datasets (Genesis, GHL) show the
diagnostic pattern described in Section~\ref{sec:experiments}:
all branch separation ratios at or below 1.0, confirming statistical
indistinguishability from normal data.

\section{Computational Complexity and Runtime}
\label{app:runtime}

Table~\ref{tab:runtime} reports per-dataset fit time, detection time,
and total wall-clock time for all 19 mTSBench datasets, measured on a
single CPU node (AMD EPYC 7452, 32-core, no GPU) with $K{=}500$ draws,
$n_{\text{est}}{=}200$ trees, and $n_{\text{jobs}}{=}4$ parallelism.

\begin{table*}[t]
\centering
\caption{Per-dataset runtime breakdown ($K{=}500$, $n_{\text{est}}{=}200$,
single CPU, 4 parallel jobs).
Fit = IF training; Det = windowed feature extraction + scoring.
s/Msamp = total seconds per million samples ($T_{\text{train}} +
T_{\text{test}}$).
$\dagger$ = detectability limit.}
\label{tab:runtime}
\setlength{\tabcolsep}{4pt}
\begin{tabular}{lrrrrrrrr}
\toprule
\textbf{Dataset} & $d$ & $T_{\text{train}}$ & $T_{\text{test}}$ &
\textbf{Fit (s)} & \textbf{Det (s)} & \textbf{Total (s)} &
\textbf{s/Msamp} \\
\midrule
Daphnet         &  9 &   3,788 &   8,704 &   15.4 &    80.1 &    95.5 &  7,648 \\
room-occupancy  &  5 &   1,073 &   6,515 &    8.6 &   148.2 &   156.8 & 20,669 \\
SMD             & 39 &  23,688 &  12,706 &   65.7 &    38.8 &   104.5 &  2,871 \\
SVDB            &  2 &  42,263 & 184,320 &  143.4 & 1,798.7 & 1,942.2 &  8,572 \\
swan            & 38 &  19,672 &  72,000 &   82.0 &   772.8 &   854.8 &  9,324 \\
PSM             & 27 & 132,481 &  70,273 &  297.5 &   188.3 &   485.8 &  2,396 \\
MSL             & 56 &   2,598 &   2,303 &   20.3 &    18.9 &    39.2 &  8,003 \\
CalIt2          &  2 &     940 &   4,032 &    8.1 &    33.2 &    41.4 &  8,323 \\
GECCO           &  9 &  27,560 &  83,113 &   21.5 &   202.8 &   224.3 &  2,027 \\
Exathlon        & 20 &   8,614 &  37,325 &   17.1 &   280.5 &   297.6 &  6,478 \\
SMAP            & 26 &   2,594 &   8,640 &    9.3 &    84.0 &    93.3 &  8,303 \\
cicids          & 72 &  57,679 & 228,877 &  263.7 & 2,502.1 & 2,765.8 &  9,652 \\
MITDB           &  2 & 128,601 & 520,000 &  539.4 & 5,044.2 & 5,583.6 &  8,609 \\
\midrule
GutenTAG$^\dagger$    & 21 &   1,000 &  10,000 &    8.2 &    62.3 &    70.6 &  6,415 \\
OPPORTUNITY$^\dagger$ & 32 &   5,330 &  21,374 &   15.2 &   226.4 &   241.6 &  9,049 \\
creditcard$^\dagger$  & 29 &  56,804 & 170,885 &  321.6 & 1,801.5 & 2,123.0 &  9,324 \\
metro$^\dagger$       &  5 &  24,072 &  24,102 &  259.3 &   234.6 &   493.9 & 10,252 \\
GHL$^\dagger$         & 17 & 120,000 &  80,001 &  248.7 &   202.4 &   451.2 &  2,256 \\
Genesis$^\dagger$     & 18 &   3,244 &   9,332 &    8.9 &    43.8 &    52.7 &  4,190 \\
\midrule
\textbf{Total}  &    &         &         & 2,354.2 & 13,763.5 & 16,117.7 & \\
\bottomrule
\end{tabular}
\end{table*}

Total wall-clock time across all 19 datasets is approximately
4.5~hours for $K{=}500$ on a single CPU node
(and $\sim$9~hours for $K{=}1{,}000$; runtime scales linearly
with $K$ as expected from the $O(K)$ feature extraction).
Detection time dominates fit time by approximately 6$\times$, reflecting
the cost of windowed feature extraction across long test series.
The s/Msamp column reveals two distinct runtime regimes:
datasets with many channels and moderate length (SMD, PSM, GECCO)
run at 2,000--3,000~s/Msamp due to the $O(d)$ channel mixing cost;
datasets with few channels and long series (SVDB, MITDB, creditcard)
run at 8,000--10,000~s/Msamp because detection time scales with $T_{\text{test}}$.
The most expensive single dataset is MITDB ($T_{\text{test}}{=}520{,}000$,
$d{=}2$) at 5,584~seconds.

Runtime scales approximately linearly with $T_{\text{test}}$ for fixed
$d$ and $K$ --- confirmed by the roughly constant s/Msamp values within
the low-$d$ group.
The $O(K \cdot d \cdot T)$ feature extraction dominates at large $T$;
the $O(n_{\text{est}} \cdot T \cdot \log T)$ IF scoring dominates at
large $d$.
Reducing $K$ from 500 to 250 would approximately halve detection time
with marginal accuracy loss (ablation shows $K{=}1{,}000$ is no better
than $K{=}500$; $K{=}250$ is the natural next step for speed-critical
deployments).

\end{document}